%% file: main.tex
\documentclass[letterpaper, 10 pt, conference]{ieeeconf}  

\IEEEoverridecommandlockouts                              

\usepackage{graphics} 
\usepackage{epsfig} 
\usepackage{amsmath} 
\usepackage{subfigure}
\usepackage{color}
\usepackage{hhline}
\usepackage{tabularx}
\usepackage{hyperref}

\graphicspath{
	{pictures/}
}

\newcommand{\myparagraph}[1]{\noindent\textbf{#1}}

\title{\LARGE \bf
Analyzing Modular CNN Architectures \\ for Joint Depth Prediction and Semantic Segmentation
}

\author{Omid Hosseini Jafari$^{*}$, Oliver Groth$^{*}$, Alexander Kirillov$^{*}$, Michael Ying Yang$^{\dagger}$ and Carsten Rother$^{*}$%
\thanks{$^{*}$Computer Vision Lab Dresden,
        Technical University Dresden, Germany
        {\tt\small http://cvlab-dresden.de/people/}}%
\thanks{$^{\dagger}$Scene Understanding Group,
				    University of Twente, Netherlands
        {\tt\small michael.yang@utwente.nl}}%
}

\begin{document}

\maketitle
\thispagestyle{empty}
\pagestyle{empty}

\begin{abstract}
This paper addresses the task of designing a modular neural network architecture that jointly solves different tasks. As an example we use the tasks of depth estimation and semantic segmentation given a single RGB image.
The main focus of this work is to analyze the cross-modality influence between depth and semantic prediction maps on their joint refinement. While most previous works solely focus on measuring improvements in accuracy, we propose a way to quantify the cross-modality influence.
We show that there is a relationship between final accuracy and cross-modality influence, although not a simple linear one. Hence a larger cross-modality influence does not necessarily translate into an improved accuracy.
We find that a beneficial balance between the cross-modality influences can be achieved by network architecture and conjecture that this relationship can be utilized to understand different network design choices.
Towards this end we propose a Convolutional Neural Network (CNN) architecture that fuses the state of the state-of-the-art results for depth estimation and semantic labeling. By balancing the cross-modality influences between depth and semantic prediction, we achieve improved results for both tasks using the NYU-Depth v2 benchmark.
\end{abstract}

\section{Introduction}
\label{sec:intro}

\begin{figure*}[h]
    \centering
    \includegraphics[width=0.7\textwidth]{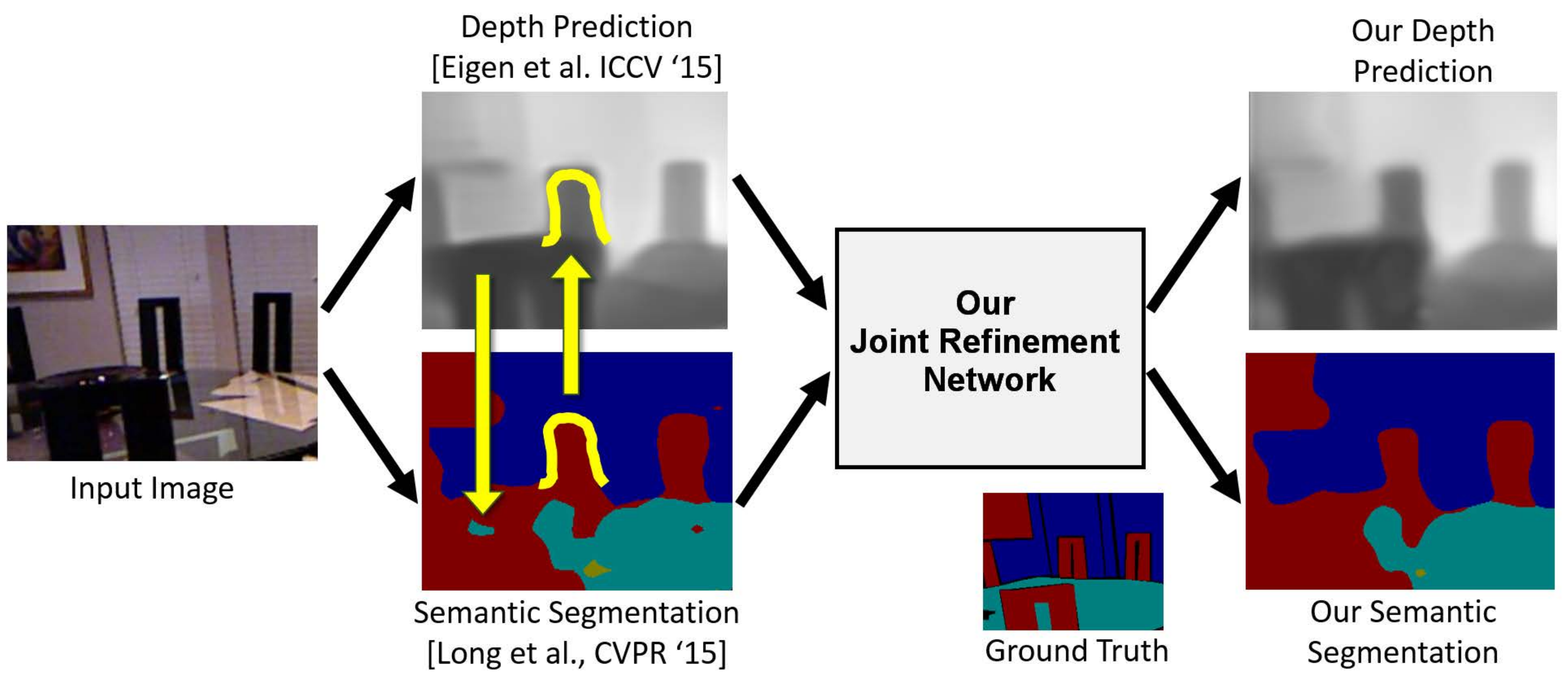}
    \caption{{\bf Example processing flow of our joint refinement network}. A single RGB image is first processed separately by two state-of-the-art neural networks for depth estimation and semantic segmentation. The two resulting predictions contain information which can mutually improve each other: (1) yellow arrow from depth to semantic segmentation means that a smooth depth map does not support an isolated region (cyan means furniture); (2) yellow arrow from semantic segmentation to depth map means that the exact shape of the chair can improve the depth outline of the chair. (3) In most areas the two modalities positively enforce each other (e.g. the vertical wall (dark blue) supports a smooth depth map. The \emph{cross-modality influences} between the two modalities are exploited by our joint refinement network, which fuses the features from the two input prediction maps and jointly processes both modalities for an overall prediction improvement.  \textit{(Best viewed in color.)}}
    \label{fig:pull-figure}
\end{figure*}

Machine Perception is an important and recurrent theme throughout the Robotics and Computer Vision community. Computer Vision has contributed a broad range of tasks to the field of perception, such as estimating physical properties from an image, e.g. depth, motion, or reflectance, as well as estimating semantic properties, e.g. labeling each pixel with a semantic class. One may argue that all of these tasks contribute to one central goal, which can be broadly described as ``holistic scene understanding''.
In the last decade a lot of research effort has focused on solving individual tasks as good as possible. While it is certainly important to gauge the limits of individual tasks, various researchers have recently raised the question of whether the next big step forward can be achieved by focusing on improving single tasks or by considering different tasks in a joint fashion, e.g. \cite{Li2012:hsu}.\footnote{See for example the recent workshop on ``Recognition meets Reconstruction'', where one aim is to solve two tasks jointly.} This question is particularly emphasized in robotics setups where the coordination of multiple tasks and consolidation of various predictions is constitutive.
In this work we focus on the question of {\it ``How to analyze and exploit the cross-modality influence between depth and semantic predictions in order to solve tasks jointly.''}.  While the idea of a ``beneficial influence between different tasks'' is not new, it has in our opinion not received enough attention. This is in contrast to other fields, such as neuroscience, psychology and machine learning.
In principle there are two different ways to consider multiple tasks in a joint framework. One option is to formalize one big, joint model. For example this can be a neural network which has a single RGB image $I$ as input and outputs a semantic segmentation $S$ and a depth labeling $D$; or a graphical model which represents $p(S,D|I)$.
While this is a popular choice and many impressive results have been achieved, e.g. \cite{NIPS2013_5198} (for depth, semantic segmentation and more) or \cite{BarronTPAMI2015} (for depth, surface reflectance and lighting), it has its drawbacks. Firstly, the models become rapidly complex and are hence rarely used in follow-up works. Secondly, it is very difficult to analyze whether there is indeed a beneficial influence between different tasks. For instance, as we will see later, a joint model may have interdependency between modalities and tasks and can in fact be considered as two separate models.
The second possible approach for solving multiple tasks jointly is to follow a modular design. In this work we pursue this option. We propose a simple modular design where individual tasks are first inferred separately and then fed into our {\it joint refinement network} (see Fig.~\ref{fig:pull-figure}). The aim of this network is to leverage a beneficial cross-modality influence between the soft (probabilistic) input modalities in order to jointly refine both task outputs. 
We show experimentally that there is indeed a relation between the cross-modality influence and an improvement in accuracy for the individual tasks. However, the relation is not a simple linear one, i.e. a larger cross-modality influence does not necessarily mean higher accuracy.

While such a modular design is not as rich as a complex joint model, it brings many advantages: 
(i) New modalities can be easily integrated. For instance a module that estimates the reflectance properties can be integrated. 
(ii) We can quantify the cross-modality influence between different modalities, as discussed in detail later. 
(iii) It is easier to train all the tasks, in contrast to a full joint model. For example, in practice we often have many training images for individual modalities but fewer training images for all modalities jointly. A joint model would have to be trained in a semi-supervised fashion in order to cope with such heterogeneous data, while in a modular architecture each module is trained with the applicable training data. 
(iv) Since in our case each module, i.e. for the individual task and the joint refinement, is realized in the form of Convolution Neural Networks (CNNs), it is possible to conduct end-to-end training.

The advantages of modular architectures are not new and indeed David Marr describes it nicely in his book (\cite{Marr1982} page 102): ``This principle [of modular design] is important because if a process is not designed in this way, a small change in one place has consequences in many other places. As a result, the process as a whole is extremely difficult to debug or to improve, whether by a human designer or in the course of natural evolution.'' 

To summarize, our {\bf main contributions} are threefold:
\begin{itemize}
\item For both tasks, semantic segmentation and depth estimation, we improve on the state-of-the-art results for the NYU-Depth v2 benchmark \cite{nyuv2-eccv12}. We achieve this by proposing a new, joint refinement network which takes as input the results of the current state-of-the-art networks for the individual tasks.
\item For modular architecture designs we propose an experimental set-up to measure the cross-modality influence quantitatively. Such experiments are well-known in neuroscience, but have not yet been used in computer vision or robotics, to the best of our knowledge  
\item We analyze different network designs with respect to their cross-modality influence and show that there is indeed a relationship between the cross-modality influences and task performances. Although not linear, this relationship can be used to understand  different design choices in network architectures.
\end{itemize}

\section{Related Work}
\label{sec:related}

A large body of work in computer vision has focused on the two separate problems of semantic segmentation and depth estimation. In the review below, we focus on techniques that specifically address multi-modal architectures or perform semantic segmentation and depth estimation from a single monocular image.

\myparagraph{Single tasks.}
Conditional Random Fields (CRFs) are popular models that have been used in both depth estimation task, e.g.~\cite{saxena-nips05,saxena-pami09,bliu-cvpr10,mliu-cvpr14,hane-cvpr15,zhou-cvpr15}, and semantic segmentation task, e.g.~\cite{Shotton:20096,LadickyRKT14}. Such approaches predominantly use hand-crafted features.
Recently, convolutional neural networks (CNNs) are driving advances in computer vision, such as for image classification \cite{krizhevsky-cnn-2012}, object detection \cite{zhang-rcnn-2014,gupta_eccv14}, recognition \cite{agrawal-eccv14,oquab-cvpr14}, semantic segmentation \cite{girshick-cvpr14,long-shelhamer-fcn-2015}, pose estimation \cite{toshev-pose-2014} and depth estimation \cite{eigen-nips14}. The success of CNNs is attributed to their ability to learn rich feature representations as opposed to hand-designed features. Eigen et al. \cite{eigen-nips14} trained multi-scale CNNs for depth map prediction from a single image. 
Liu et al. \cite{Liu16pami} propose deep convolutional neural fields for depth estimation, where a CRF is used to explicitly model the relations of neighboring superpixels, and the potentials are learned in a unified CNN framework.
Eigen and Fergus \cite{eigen-iccv15} extend their previous method \cite{eigen-nips14} to predict depth, surface normals and semantic labels sequentially with a common multi-scale CNN. A number of recent approaches, including recurrent CNNs (R-CNNs) \cite{pinheiro-icml14} and fully convolutional networks (FCN) \cite{long-shelhamer-fcn-2015} have shown a significant boost in accuracy by adapting state-of-the-art CNN-based image classifiers to the semantic segmentation problem. Pinheiro and Collobert \cite{pinheiro-icml14} present a feed-forward approach for scene labeling based on an R-CNN. The system is trained in an end-to-end manner over raw pixels and models complex spatial dependencies with low computational cost. FCNs \cite{long-shelhamer-fcn-2015} address the coarse-graining effect of the CNN by upsampling the feature maps in deconvolution layers and combining fine-grained and coarse-grained features during prediction. 

\myparagraph{Joint models.}
Joint models of multiple tasks have been exploited in the computer vision literature to a certain extent,~e.g.~joint image segmentation and stereo reconstruction~\cite{Bleyer2011,Guillemaut2011,Ladicky2012}, joint object detection and semantic segmentation~\cite{Yao2012}, joint instance segmentation and depth ordering \cite{zhang*iccv15}, as well as joint intrinsic image, objects, and attributes estimation \cite{NIPS2013_5198}.
However, joint semantic segmentation and depth estimation from a single image has been rarely addressed, with  a few exceptions \cite{ladicky-cvpr14,wang-cvpr15}.
These works explicitly reason about class segmentation as well as depth estimation from a single image. Ladicky \textit{et al.} \cite{ladicky-cvpr14} jointly trained a canonical classifier considering both the loss from semantic and depth labels of the objects. However, they use local regions with hand-crafted features for prediction, which is only able to generate very coarse depth and semantic maps. Wang et al. \cite{wang-cvpr15} formulate the joint inference problem in a two-layer Hierarchical Conditional Random Field (HCRF). The unary potentials in the bottom layer are pixel-wise depth values and semantic labels, which are predicted by a CNN trained globally using the full image, while the unary potentials in the upper layer are region-wise depth and semantic maps which come from another CNN-based regressor trained on local regions. The mutual interactions between depth and semantic information are captured through the joint training of the CNNs and are further enforced in the joint inference of HCRF. They consider an alternating optimization strategy by minimizing one, fixing the other.
In contrast, our model performs full joint inference.
%

\myparagraph{Multi-modal learning and representation.}
Many different communities have addressed the problem of multi-modal learning and representation, such as 
machine learning \cite{icml11_ngiam,NIPS2012_4683,ChandarKLR15},
human-computer interaction \cite{obrenovic04,JaimesS07},
and neuroscience \cite{vonKriegstein08,Schall13}.
In \cite{icml11_ngiam}, the authors present a series of tasks for multi-modal learning and show how to train deep networks that learn features to address these tasks.
In particular, they demonstrate cross modality feature learning, where better features for downstream classification tasks are learned from a video if both audio and video signals are present during the feature learning stage.
While \cite{icml11_ngiam} deals with an unsupervised feature learning, our approach uses supervised learning.
Furthermore, unlike \cite{icml11_ngiam} we perform an analysis on the effect of different network architectures on the cross-modality influence.
Similarly, in the neuroscience community, the authors of \cite{vonKriegstein08} investigated the influence of the \emph{face-benefit} in speech and speaker recognition. Apparently, people who have heard the voice and seen the face of a speaker during training time are more likely to recognize both the speaker and the spoken words from recorded audio only during test time. Additionally, \cite{Schall13} revisited the \emph{face-benefit} experiment and showed a joint audio-visual processing by the brain for the classification task, indicating a joint feature representation is key to superior performance.
Canonical correlation analysis (CCA) \cite{hotelling36} is the de-facto approach for learning a common representation of two different modalities (so-called views) in the machine learning literature.
Deep CCA, a deep learning version of CCA, is introduced in \cite{icml2013_andrew13}. It aims at learning a complex non-linear transformation of two views such that the resulting representation is highly correlated. It can be considered as a non-linear extension of the linear CCA.
In this paper, we quantify the cross-modality influence in an \emph{influence number} which characterizes the magnitude of the contribution of a particular modality to the final model performance, dependent on the model architecture.

\input{synergy_network}

\input{experiments}
\section{Conclusions}
\label{sec:conclusions}
Inspired by work in neuroscience we have introduced a systematic way to measure the cross-modality influence present in our JRN networks. By doing so, we were able to identify a network which achieves a measurable influence between modalities, has an overall good performance compared to other JRN networks, and is consistently better than the state-of-the-art input modalities.


\section*{Acknowledgements}
This project has received funding from the European Research Council (ERC) under the European Union’s Horizon 2020 research and innovation program (grant agreement No 647769), and German Research Foundation (DFG) YA351/2-1. The authors gratefully acknowledge the support.


\bibliographystyle{IEEEtran}
\input{main.bbl}

\end{document}

%% file: synergy_network.tex
\section{Joint Refinement Network}
\label{sec:model}

In this section we present the details of the CNN architecture which we used to predict jointly the depth map and the semantic labeling. We also discuss different architectural design decisions and their relation to the cross-modality influence between two modalities.

\subsection{Network Architecture}

\begin{figure}[h!]
    \centering
    \includegraphics[width=\columnwidth]{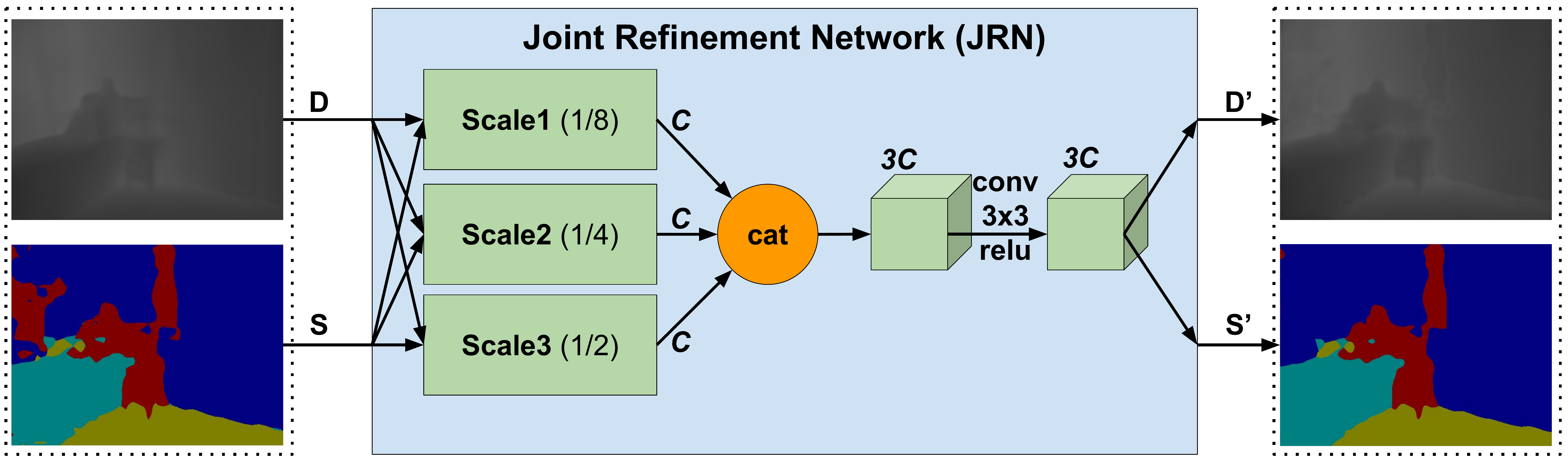}
    \caption{\textbf{Overall Network.} JRN receives as input the predictions of two independent single modalities: Depth and semantic labeling. Inspired by \cite{eigen-iccv15} the inputs are considered at different scales ($1/8$, $1/4$ and $1/2$ of the total image resolution) in order to capture different levels of details. $C$ is the number of output feature channels from each scale branch (see \ref{sec:syn-net-variants}). After processing the three scale branches, the computed features are concatenated, convolved and then mapped to the two respective output maps. }
    \label{fig:synergy-net}
\end{figure}

Our network decomposes into two parts: (i) independent \emph{single-modality} models that output predictions for each modality separately and (ii) our joint refinement network (JRN) that takes as input these prediction maps and outputs refined predictions of all modalities. Our model does not have any constraints on the choice of \emph{single-modality} models.
In order to capture dependencies on different scales, we employ a multi-scale architecture for JRN, as illustrated in Fig.~\ref{fig:synergy-net}. It has three \emph{scale-branches} {\tt Scale1}, {\tt Scale2} and {\tt Scale3} that work with different scales of the input and have the same architecture, described in Fig.~\ref{fig:scale-branch}. On each scale, 20-dimensional features are extracted by performing $3 \times 3$ convolutions on each input modality. After each convolutional layer, a ReLU non-linearity is used.
In the next section, we consider different architecture designs for the combination.

\subsection{JRN Variants}
\label{sec:syn-net-variants}

\begin{figure}
    \centering
    \includegraphics[width=\columnwidth]{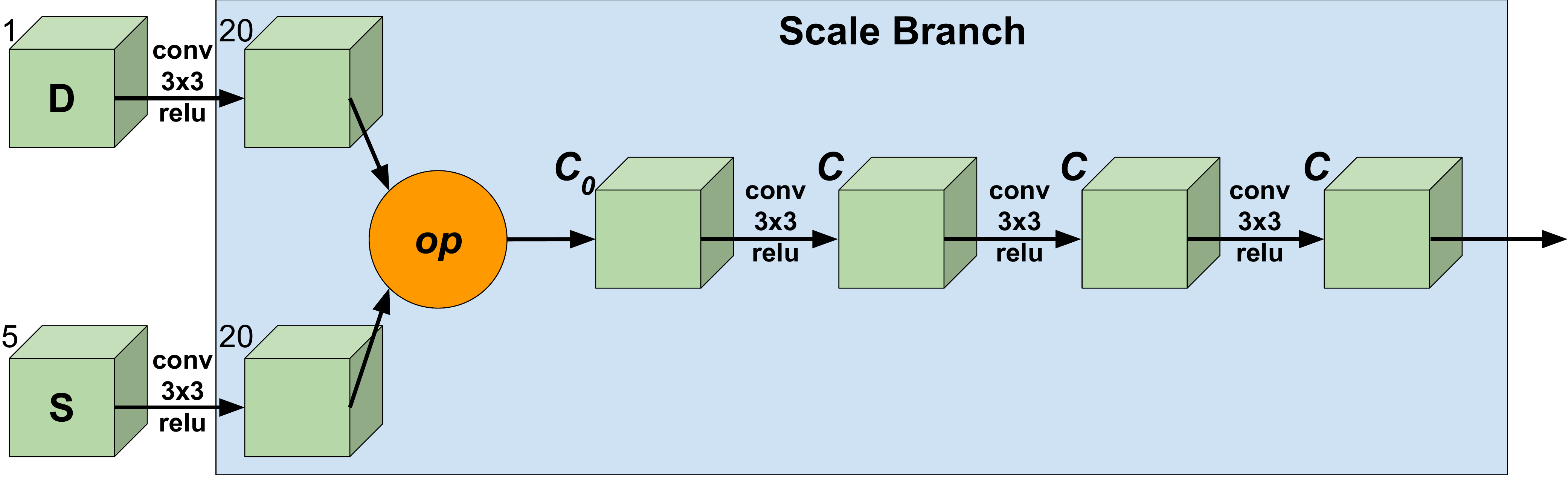}
    \caption{\textbf{A Scale Branch.} On each scale there are first 20-dimensional feature vectors extracted by performing 3x3 convolutions on each input modality. Immediately thereafter these modality features are fused by operation \textit{\textbf{op}}. We also consider the number of channels $\boldsymbol{C_0}$ after the fusion operation as a network design variable, which affects the cross-modality influence. The subsequent channel number \textit{\textbf{C}} is 60 by default (\texttt{Cat60}, \texttt{Sum60}). Based on \textbf{\textit{op}}, $\boldsymbol{C_0}$ and \textit{\textbf{C}} we design five different network architectures and analyze their properties in Sec. \ref{sec:comparison-of-results}. }
    \label{fig:scale-branch}
\end{figure}

\myparagraph{Concatenation.}
In the \emph{concatenation} architecture we have chosen concatenation operations for \textit{\textbf{op}} in all scale branches consistently (cf.~Fig.~\ref{fig:scale-branch}). 
Therefore, the number of channels $\boldsymbol{C_0}$ after concatenation has doubled to $40$. 
The simple concatenation of the single-modality features aims at achieving cross modality features because the single-modality features are subsequently convolved together. We call this variant \texttt{Cat60}. Since concatenation is the weakest form of our possible feature interactions, we subsequently refer to this model as featuring a \emph{loose computation}.

\myparagraph{Summation.}
The \emph{summation} architecture features an element-wise sum as fusion operation \textit{\textbf{op}} on each scale. Since the channels are summed up during fusion, the number of subsequent channels $\boldsymbol{C_0}$ is equal to $20$. We call this variant \texttt{Sum60}. In contrast to the \texttt{Cat60} architecture this network forces a joint feature representation. We refer to this architecture as having a \emph{coupled computation} later on.

\myparagraph{Channel Squeezing.} Besides summation of input features we also consider the number of channels used after fusion as an influencing factor on the cross-modality influence and the overall network performance. We hypothesize that fewer channels would generally enforce network to take into account both modalities more during training of the convolutional layers. Directly after the concatenation (\textbf{\textit{op}}) of the input features, which yields a 40-channel feature, the number of channels \textit{\textbf{C}} can be 10, 5 or 1. This gives us three more network variations \texttt{Cat10}, \texttt{Cat5} and \texttt{Cat1}. We still classify the \texttt{Cat10} network as \emph{loose computation} whereas the \texttt{Cat5} and \texttt{Cat1} networks featuring \emph{coupled computation}.

\subsection{JRN Training}

We define $D'$ and $S'$ as the depth and semantic label prediction maps and $D^*$ and $S^*$ as the respective ground truth maps.
$D'$ and $D^*$ are maps assigning a depth value in the range of $[0.0,10.0]$ meters to every pixel. $S'$ and $S^*$ are $k$-channel maps, assigning a probability distribution over $k$ semantic classes to each pixel.
We restrict the loss computation to $n$ valid pixels where we have both a depth value and a semantic label as ground truth.
The loss function for JRN is a simple summation of single-task losses:
\begingroup\makeatletter\def\f@size{6}\check@mathfonts
\def\maketag@@@#1{\hbox{\m@th\large\normalfont#1}}%
\begin{align}
    L_{joint}(D',S',D^*,S^*) = \underbrace{\frac{1}{n}\sum_{i}\frac{(D'_i-D_i^*)^2}{D_i^*}}_{L_{depth}} -\underbrace{\frac{1}{n}\sum_{i}S_i^*\log(S'_i)}_{L_{semantic}}
\end{align}
\endgroup

The depth loss $L_{depth}$ is a relative quadratic distance between prediction and ground truth map.
For the semantic loss $L_{semantic}$ we use the cross-entropy loss, where $S'_i = \exp[z_i] / \sum_{s}{\exp[z_{i,s}]}$ is the class prediction at pixel $i$ given the semantic output slice $z$ of the last convolutional layer of JRN.
During training we keep the single-task networks fixed and use their predictions as inputs to JRN. In the future we plan to also perform end-to-end training of all networks, single task networks and JRN. 
The internal weights of JRN are initialized randomly. We train JRN jointly on both tasks with the standard NYU-Depth v2 \cite{nyuv2-eccv12} train-test split.

\subsection{Quantifying the Cross-Modality Influence}
\label{sec:synergy-test}

\begin{figure}
    \centering
    \includegraphics[width=\columnwidth]{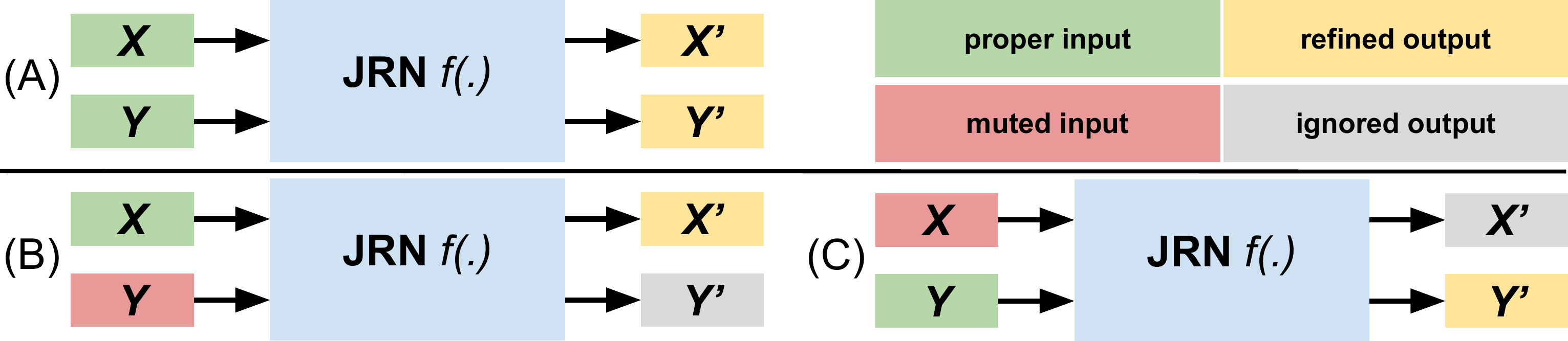}
    \caption{\textbf{Cross-Modality Influence Test.} In order to quantify the cross-modality influence numbers we need to consider three different inference setups: (A) Predicting both outputs $X'$ and $Y'$ from two proper input maps $X$ and $Y$: $(X',Y') = f(X,Y)$; (B) Muting the input channel $Y$ and computing $X'$ and $Y'$: $(X',Y') = f(X)$; and (C) Muting the input channel $X$ and computing $X'$ and $Y'$: $(X',Y') = f(Y)$.}
    \label{fig:synergy-test}
\end{figure}

Inspired by the ``face benefit'' experiment in \cite{vonKriegstein08} we propose an evaluation proxy to measure the influence of a modality on the final model performance. During training time the JRN is trained to jointly predict two outputs $X'$ and $Y'$ from two input modalities $X$ and $Y$. During inference time we consider three different measurement setups as explained in Fig. \ref{fig:synergy-test}. The \emph{performance} of the JRN predicts a particular modality $X'$ which is measured by a function $A_X$ (e.g. ``mean IOU'' for $X$ being semantic labeling or ``rms(linear)'' for $X$ being a depth map).

\noindent
The cross-modality influences between the modalities are \emph{directional} and dependent on a particular JRN architecture (represented by its transformation function $f(.)$) as well as two performance functions $A_X$ and $A_Y$.
The cross-modality influence of input modality $Y$ on the performance of the prediction of $X'$ measured under $A_X$ is defined as:

\begin{align}
    \omega_{Y \rightarrow X'} = A_X(f(X,Y)) - A_X(f(X)) .
\end{align}

Consequently the complementary influence from $X$ to $Y'$ is defined as:

\begin{align}
    \omega_{X \rightarrow Y'} = A_Y(f(X,Y)) - A_Y(f(Y)) .
\end{align}

Note that those influences are \emph{not necessarily symmetric}.
We compute the influence values in the setup of a joint semantic segmentation ($X$) and depth estimation ($Y$) in Sec. \ref{sec:perf-syn-rel} and analyze the relationship between the cross-modality influences and the model performance.

%% file: experiments.tex
\section{Experiments}
\label{sec:experiments}

In this section, we introduce the dataset, and then describe details of our implementation.
After that we present a quantitative and qualitative comparison. We conclude with a cross-modality influence analysis.

\subsection{Experimental setup}

\myparagraph{Dataset.}
We evaluate our proposed method on the NYU-Depth V2 dataset \cite{nyuv2-eccv12}. This dataset consists of $1449$ RGBD images of indoor scenes, among which $795$ are used for training and $654$ for test (we use the standard train-test split provided with the dataset).
Following \cite{wang-cvpr15}, we also map the semantic labels into five categories conveying strong geometric properties, i.e.~{Ground, Vertical, Ceiling, Furniture and Object}, as it is shown in Fig. \ref{fig:results-qualitative}.

\myparagraph{Implementation details.}
We use two state-of-the-art single modality CNN models for providing the input depth-map and semantic segmentation prediction map. For depth input, we use the inference code with pretrained model of Eigen \textit{et al.} \cite{eigen-iccv15} which is publicly available\footnote{\url{http://www.cs.nyu.edu/~deigen/dnl/}}. For semantic segmentation prediction maps, we use the FCN model of Long \textit{et al.} \cite{long-shelhamer-fcn-2015}. 

We implement our network in Caffe framework \cite{jia2014caffe}. We train joint refinement networks (\texttt{Cat60}, \texttt{Cat10}, \texttt{Cat5}, \texttt{Cat1} and \texttt{Sum60}) with 795 training images from NYU-Depth V2 dataset \cite{nyuv2-eccv12} using SGD solver with batches of size one. The learning rate is 0.001 for all the convolutional layers and the momentum is 0.9. The global scale of the learning rate is tuned to a factor of 5.  Depending on the different architectures and the number of channels in scale branches, training JRN took 5 to 6 hours using an NVidia GTX Titan X. We pass the absolute depth maps and the semantic prediction maps to JRN.

\myparagraph{Evaluation metrics.}
To evaluate the semantic segmentation, we take Intersection over Union (IOU) and pixel accuracy percentage as metrics. 
For the depth estimation task we use several measures, which are also commonly used in prior works \cite{mliu-cvpr14,eigen-nips14}. Given the predicted depth value of a pixel $d_i$ and the ground truth depth $d_i^*$, the evaluation metrics are:
 
 \begin{itemize}
     \item Abs relative error (rel):
     $\frac{1}{N} \sum_i \frac{|d_i^{*} - d_i|}{d_i^{*}} $;
     \item Squared relative error (rel(sqr)):
     $\frac{1}{N} \sum_i \frac{|d_i^{*} - d_i|^2}{d_i^{*}} $;
     \item Average $\log_{10}$ error (log10): 
     $\frac{1}{N} \sum_i | \log_{10}d_i^{*} - \log_{10}d_i|$;
     \item Root mean squared error (rms(linear)):
     $\sqrt{\frac{1}{N} \sum_i (d_i^{*} - d_i)^2}$;
     \item Root mean squared error (rms(log)):
     $\sqrt{\frac{1}{N} \sum_i | \log d_i^* - \log d_i |^2}$;
     \item Accuracy with threshold $thr$:
     percentage ($\%$) of $d_i \; \; s.t. \max (\frac{d_i^{*}}{d_i}, \frac{d_i}{d_i^*}) = \delta < thr$, where $thr \in \{1.25, 1.25^2, 1.25^3\}$;
 \end{itemize}

\subsection{Comparison of Results}
\label{sec:comparison-of-results}

We first compare our five different JRN architectures described in Sec. \ref{sec:syn-net-variants} with each other (see Table \ref{tab:synergy-param}).
We observe that none of the networks consistently outperforms all others in all metrics.
However, the \texttt{Sum60} network is nearly the best for all metrics. Hence we chose it for comparison with other models from related work.

\begin{table*}
\center
\caption{\textbf{Comparison of different JRN architectures.}
We compare our different JRN networks with each other and also with our input single modality networks for depth \cite{eigen-iccv15} and semantic segmentation \cite{long-shelhamer-fcn-2015}. Best results are shown in bold.}
\resizebox{\linewidth}{!} {
\begin{tabular}{ | l |  c  c c c  c | c  c  c | c c |}
\hline
&\multicolumn{5}{c|}{Error (Depth)} &\multicolumn{3}{c|}{Accuracy (Depth)}  &\multicolumn{2}{c|}{Accuracy (Seg.)} \\
&\multicolumn{5}{c|}{(lower is better)} &\multicolumn{3}{c|}{(higher is better)} &\multicolumn{2}{c|}{(higher is better)} \\
\cline{2-9}
&rel & rel(sqr)&log10 &rms(linear)& rms(log) &$\delta < 1.25$ &$\delta < 1.25^2$ &$\delta < 1.25^3$ & \textbf{Mean IOU} & \textbf{Pix.Acc.}  \\
\hline
\hline
Input \cite{eigen-iccv15}\&\cite{long-shelhamer-fcn-2015}  &0.158 & 0.125 & 0.070 & 0.687 & 0.221 & 0.751 & 0.946 & 0.987 & 53.284  & 72.268 \\
Cat60 & 0.158 & 0.124 & 0.0686 & 0.678 & 0.218 & 0.760 & 0.947 & 0.987 & \textbf{54.206} & 72.957 \\
Sum60 &\textbf{0.157} & \textbf{0.123} & \textbf{0.068} & 0.673 & \textbf{0.216} & \textbf{0.762}&\textbf{0.948}&\textbf{0.988} &
54.184 & \textbf{72.967} \\
Cat10 & 0.158 & 0.125 & 0.069 & 0.681 & 0.219 & 0.756 & 0.946 & 0.987 & 54.080  & 72.953\\
Cat5 & 0.160 & 0.125 & 0.068 & 0.670 & 0.218 & \textbf{0.762} & 0.946 & 0.986  & 54.120  & 72.952 \\
Cat1 & 0.161 & 0.126 & 0.069 & \textbf{0.669} & 0.219 & 0.759 & 0.946 & 0.987 & 53.989 & 72.864\\
\hline
\end{tabular}
}
\label{tab:synergy-param}
\end{table*}

\begin{table*}
\center
\caption{\textbf{Depth comparison.} Baseline comparisons of depth estimation on the NYU-Depth v2 dataset.
Our method outperforms state-of-the-art methods.}
\resizebox{0.98\linewidth}{!} {
\begin{tabular}{ | l |  c  c c c  c | c  c  c |}
\hline
&\multicolumn{5}{c|}{Error (lower is better)} &\multicolumn{3}{c|}{Accuracy (higher is better)} \\
\cline{2-9}
&rel & rel(sqr)&log10 &rms(linear)& rms(log) &$\delta < 1.25$ &$\delta < 1.25^2$ &$\delta < 1.25^3$  \\
\hline
\hline
Eigen et al. \cite{eigen-nips14} & 0.215 & 0.212 & - & 0.907 & 0.285 & 0.611 & 0.887 & 0.971 \\
Joint HCRF \cite{wang-cvpr15} & 0.220 & 0.210 & 0.094 & 0.745 & 0.262 & 0.605 & 0.890 & 0.970 \\
Eigen \& Fergus \cite{eigen-iccv15} &0.158 & 0.125 & 0.070 & 0.687 & 0.221 & 0.751 & 0.946 & 0.987 \\
Liu et al. \cite{Liu16pami} & 0.213 & - & 0.087 & 0.759 & - & 0.650 & 0.906 & 0.976 \\
Ours (Sum60) &\textbf{0.157} & \textbf{0.123} & \textbf{0.068} & \textbf{0.673} & \textbf{0.216} & \textbf{0.762} & \textbf{0.948} & \textbf{0.988} \\
\hline
\end{tabular}
}
\label{tab:depth-comparison}
\end{table*}

For \textit{depth estimation}, we compare our results with four most recent methods,~i.e. Eigen et al. \cite{eigen-nips14}, Joint HCRF \cite{wang-cvpr15}, Eigen \& Fergus \cite{eigen-iccv15}, and Liu et al. \cite{Liu16pami}.
Table~\ref{tab:depth-comparison} shows the quantitative results from all the algorithms. 
Our JRN network consistently outperforms all the state-of-the art algorithms in all metrics.
The main difference to the models of Eigen \textit{et al.} \cite{eigen-nips14}, Eigen \& Fergus \cite{eigen-iccv15} and Liu et al. \cite{Liu16pami} is that they only deal with the depth estimation task, and hence cannot exploit cross-modality influence.
However, the Joint HCRF \cite{wang-cvpr15} also jointly predicts a depth map and semantic labeling, yet our model outperforms theirs by a large margin both in depth estimation (8.7\% rel(sqr) decrease) and in semantic segmentation (10\% mean IOU increase, see Table~\ref{tab:results-semantic-c5}). We evaluate the published predicted depth maps from Eigen \& Fergus \cite{eigen-iccv15} and Eigen et al. \cite{eigen-nips14} with our evaluation script, and for Eigen et al. \cite{eigen-nips14} we obtain the same reported numbers. 
However, we do not obtain the same numbers for \cite{eigen-iccv15} as reported. Our goal in this work is to improve the performance of the input predictions. Therefore, this comparison is fair since we use the same evaluation script for the input and the output of our network.

\begin{table*}
    \centering
    \caption{\textbf{Semantic segmentation comparison.} Class-wise results reporting mean IOU and pixel-wise accuracy for semantic segmentation on NYU-Depth v2 with five classes. Best results are shown in bold.}
\resizebox{0.98\linewidth}{!} {
    \begin{tabular}{|l|c|c|c|c|c|c|c|} 
    \hline
            & \textbf{Ground} & \textbf{Vertical} & \textbf{Ceiling} & \textbf{Furniture} & \textbf{Object} & \textbf{Mean IOU} & \textbf{Pix.Acc.}\\
            \hline
        Semantic HCRF \cite{wang-cvpr15} & 61.84 & 66.344 & 15.977 & 26.291 & 43.121 & 42.715 & 69.351\\
        Joint HCRF \cite{wang-cvpr15} & 63.791 & 66.154 & 20.033 & 25.399 & 45.624 & 44.2 & 70.287\\
        FCN16s NYU-5 \cite{long-shelhamer-fcn-2015} & 66.578 & 67.354 & 46.351 & 35.71 & 50.429 & 53.284 & 72.268 \\ 
        Ours (Sum60) & \textbf{67.87} & \textbf{68.707} & \textbf{48.166} & \textbf{35.82} & \textbf{50.770} & \textbf{54.267} & \textbf{73.035}\\
        \hline
    \end{tabular}
    }
    \label{tab:results-semantic-c5}
\end{table*}

For \textit{semantic segmentation}, we compare two recent methods: Our baseline FCN \cite{long-shelhamer-fcn-2015} and Joint HCRF \cite{wang-cvpr15}.
Results are shown in Table~\ref{tab:results-semantic-c5}. 
We outperform the other methods for all five classes.
Compared with the baseline FCN \cite{long-shelhamer-fcn-2015} our method is 1\% better in mean IOU.

\begin{figure*}
    \centering
    \includegraphics[width=0.95\textwidth]{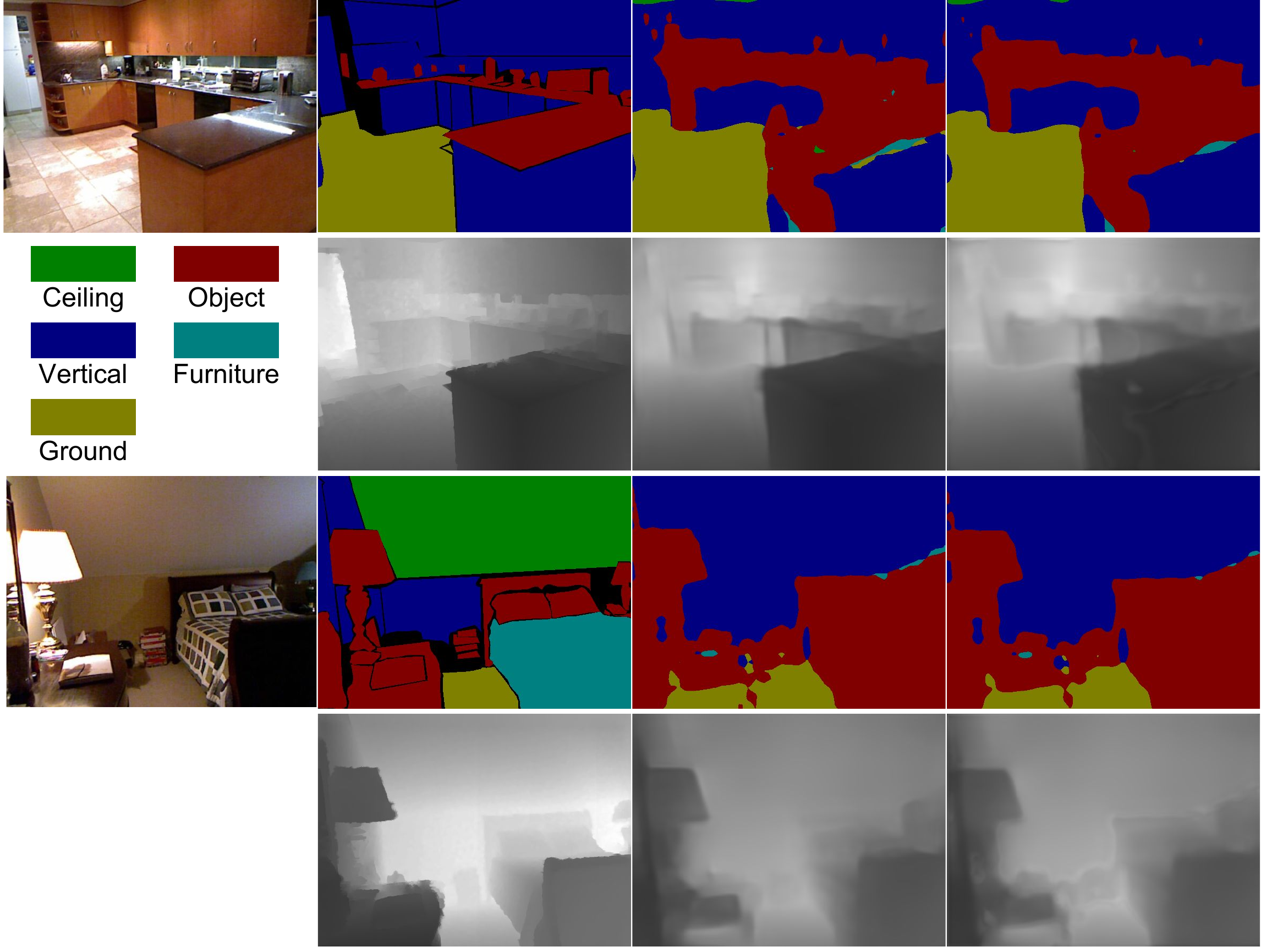}
    \caption[]{
     Two qualitative results of our \texttt{Sum60} network compared with the input (i.e state-of-the-art). Each top row (from left to right): Image, ground truth semantic labeling, result of \cite{long-shelhamer-fcn-2015} and our result. Each bottom row (from left to right): Ground truth depth map, result of \cite{eigen-iccv15} and our result. The first example depicts an improvement in semantic labeling, where ground label (yellow) has been removed (next to object label (red)). In the second example, the depth edges of the upper bed frame are better recovered in our result (best viewed zoomed-in). Please note that our results are smooth and follow edges in the input image, despite having no explicit CRF model.  
    }
    \label{fig:results-qualitative}
\end{figure*}

Qualitative results of both tasks are shown in Fig.~\ref{fig:results-qualitative}. 
Even though our method does not use superpixels or any explicit CRF model, it tends to produce large homogeneously labeled regions.

\subsection{Performance Cross-Modality Influence Analysis}
\label{sec:perf-syn-rel}

\begin{figure*}
    \centering
    \includegraphics[width=\textwidth]{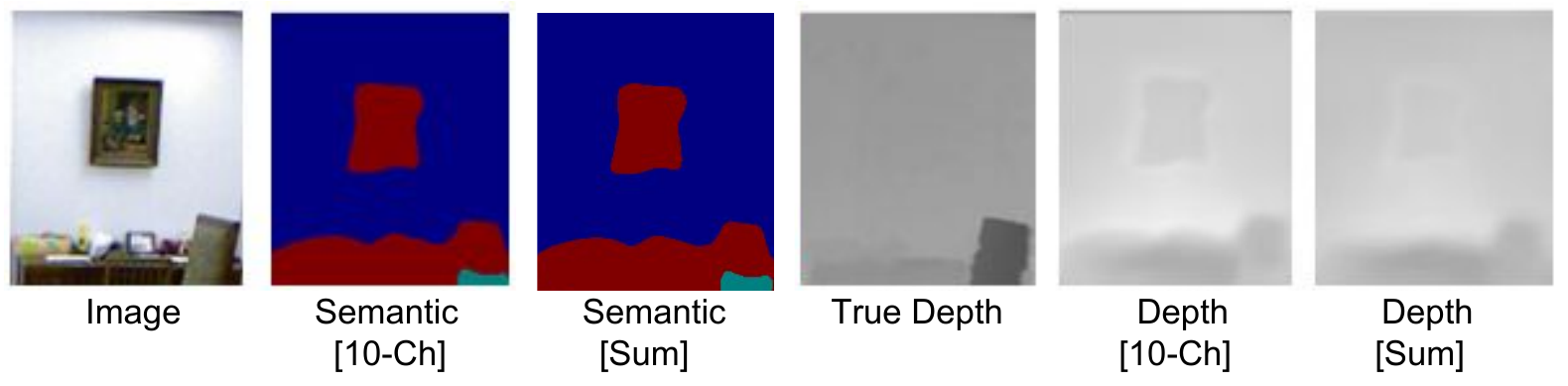}
    \caption{{\bf Negative influence.} From left to right: Input image, our semantic labeling of the \texttt{Cat10} network, semantic labeling of the \texttt{Sum60} network, ground truth depth, result of our \texttt{Cat10} network, result of our \texttt{Sum60} network. Despite the \texttt{Cat10} network having a higher cross-modality influence number $\omega_{S \rightarrow D'}$ than the \texttt{Sum60} network (cf. Fig. \ref{fig:synergy-plots} top right), the respective depth accuracy (-rel(sqr)) of the \texttt{Cat10} network is lower. This is visible in the image where the picture frame has received a wrong depth in the \texttt{Cat10} network result, compared to the \texttt{Sum60} network result. (Image crops shown for visualization purpose.)}
    \label{fig:syneffect1}
\end{figure*}

\begin{figure}
    \centering
    \includegraphics[width=0.45\columnwidth]{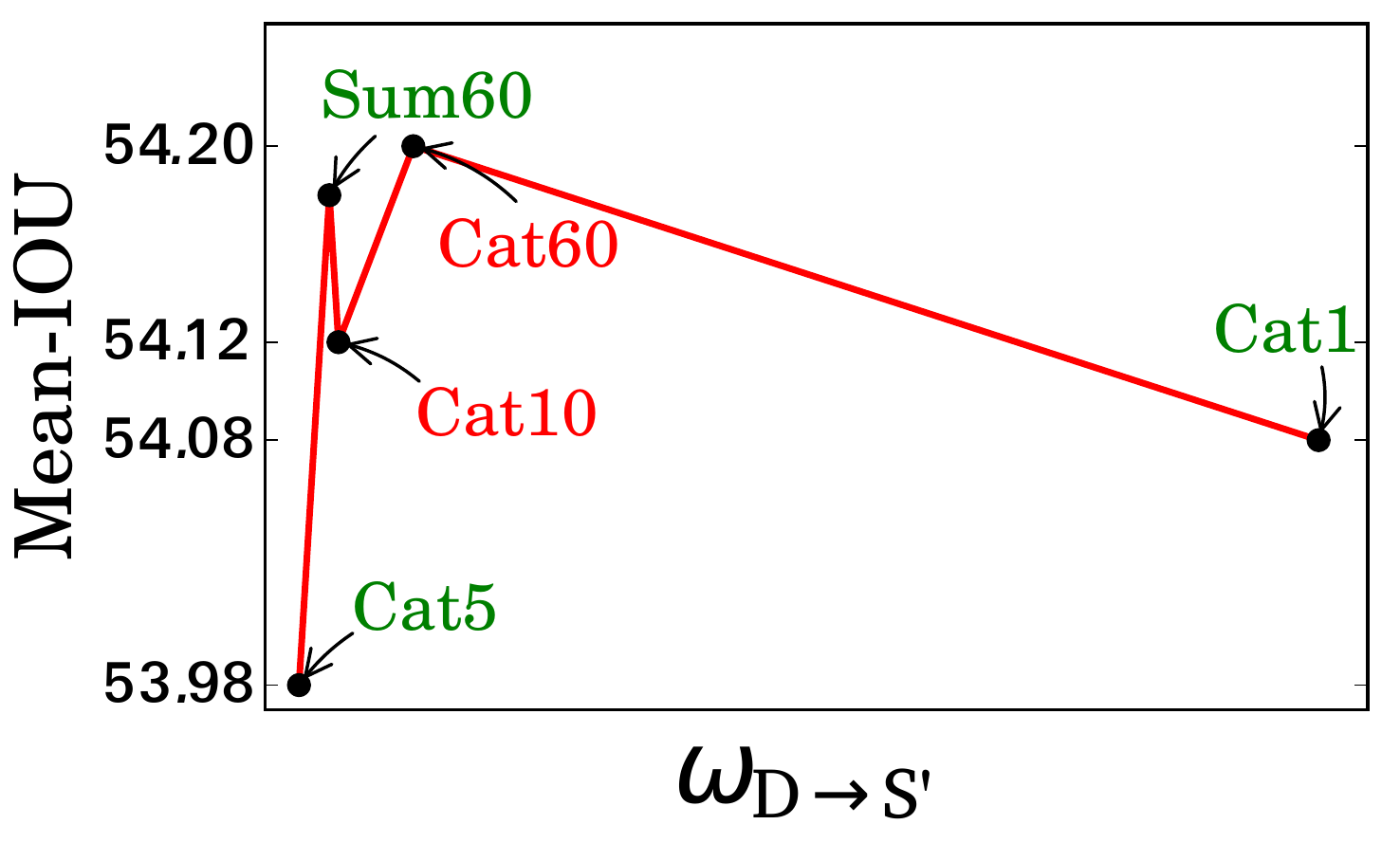}
    \includegraphics[width=0.45\columnwidth]{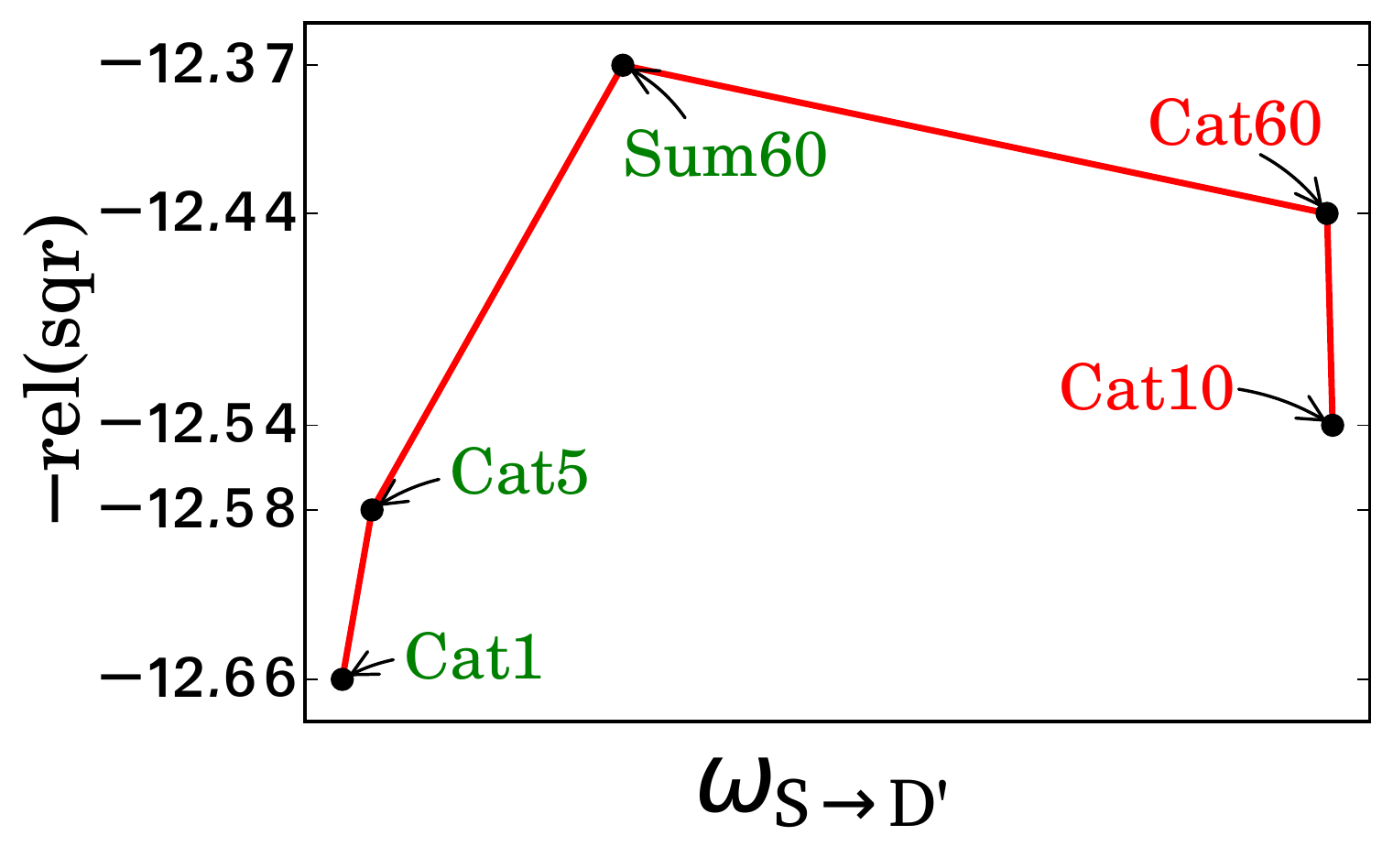}
    \quad
    \begin{tabular}[b]{|c|c|c|}
        \hline
         & $\omega_{D \rightarrow S'}$ & $\omega_{S \rightarrow D'}$\\ 
         \hline
         Cat60   & 0.94 & 4.9  \\
         Cat10 & 0.39 & 4.93 \\
         Cat5  & 0.10 & -0.24\\
         Sum  & 0.32 & 1.11 \\
         Cat1  & 7.59 & -0.40\\
        \hline
    \end{tabular}
    \caption{\textbf{Performance vs. cross-modality influence plots for all models.} We use the mean IOU measure for semantic labels ($A_S$), and $-$rel(sqr)$*100$ for depth ($A_D$). The right table shows all influence numbers $\omega_{D \rightarrow S'}$ and $\omega_{S \rightarrow D'}$ for all models. The top figures are the respective performance-influence plot. Both plots exhibit a peak where the optimal trade-off between cross-modality influence and evaluated performance is achieved. We see that the \texttt{Sum60} and \texttt{Cat60} models are at the peaks of the respective plots. We colored models in red which feature \emph{loose computation} and in green which feature \emph{coupled computation} (see Sec. \ref{sec:syn-net-variants}). This supports the idea that the cross-modality influence number can facilitate the systematic exploration of network architectures.}
    \label{fig:synergy-plots}
\end{figure}

We compute the cross-modality influence for all five JRN networks by looking at the relation between the cross-modality influence numbers and the performance in the respective modalities (see Fig. \ref{fig:synergy-plots}).
We observe that there is no linear relationship between cross-modality influence and performance but they rather lie within an area which is upper-bounded by a concave curve.
This means that a larger influence between modalities does not guarantee better performance in the respective metric.
Indeed a large negative effect can hamper performance (see Fig. \ref{fig:syneffect1}).

\begin{figure}[h!]
    \centering
    \includegraphics[width=\columnwidth]{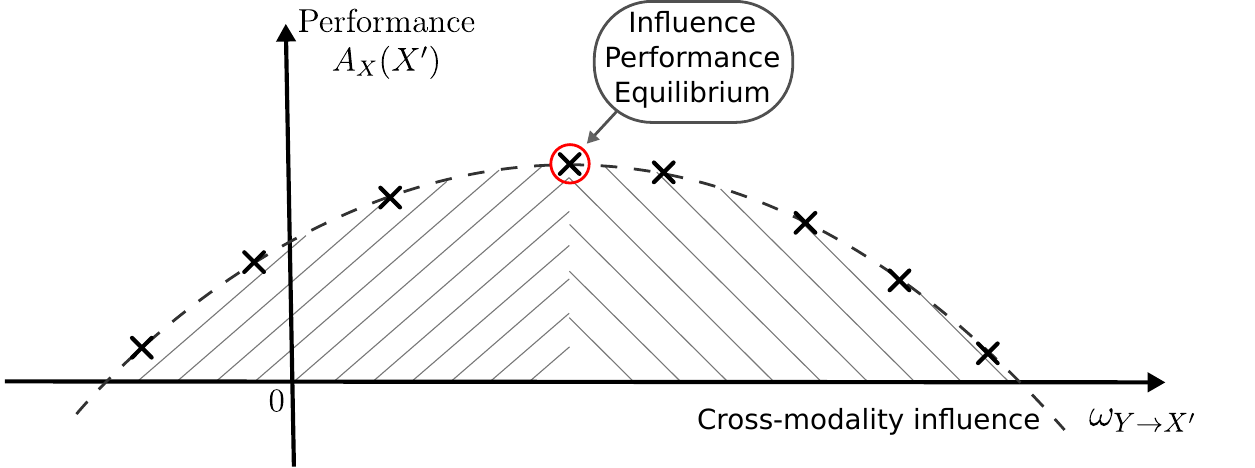}
    \caption{\textbf{The performance vs. cross-modality influence curve.} For each modality pair $(X,Y)$ and in each modality influence direction $Y \rightarrow X'$ and $X \rightarrow Y'$ the relationship between the magnitude of the cross-modality influence and the prediction performance needs to be balanced into an equilibrium. This cross-modality influence analysis will be helpful when designing models which should operate on multiple modalities and carry out joint predictions.}
    \label{fig:acc-syn-curve}
\end{figure}

Based on our findings about cross-modality influence, we hypothesize that the relationship between cross-modality influence and performance can be generalized into a plot which is sketched in Fig. \ref{fig:acc-syn-curve}.
The cross-modality influence arises from certain model design decisions, as well as from modality combinations for a particular end-task. For example, we have seen in our experiments that moderately transferring shapes and class-wise depth priors from the semantic map into the depth map can help improving depth estimation (see Fig. \ref{fig:results-qualitative} bottom). However, the cross-modality influence $\omega_{S \rightarrow D'}$ can also be too strong (large positive influence number) which causes a decrease in performance. For example, shapes from semantic segmentation can cause halos and artifacts in the depth map (see Fig. \ref{fig:syneffect1}).
We conclude that inspecting performance vs. cross-modality influence plots is a useful way to find appropriate modular architectures. Furthermore, these plots may help identifying complementary modalities to further enhance the cross-modality influence.

